\documentclass[runningheads]{llncs}
\usepackage[T1]{fontenc}
\usepackage{tabularx}
\usepackage{array}
\usepackage{booktabs}
\usepackage{standalone}
\usepackage{amsmath}
\usepackage{float}
\usepackage{multirow}
\usepackage{graphicx}
\usepackage{pgfplots}
\usepackage{tikz}

\usepackage[backend=biber,style=lncs,maxnames=10,url=false,minnames=1]{biblatex}
\usetikzlibrary{positioning, arrows.meta, shapes, backgrounds}
\usetikzlibrary{fit}
\usetikzlibrary{backgrounds}
\usetikzlibrary{calc}

\def\horiDist{0.6} 
\def\vertDist{1.7}
\def\promptEncExtraDist{0.65}
\def\blockHeight{3cm}
\def\imgSize{3cm}
\def\imgDist{0.9cm}
\def\attnXShift{0.25cm}

\tikzset{
  encoder/.style={draw, fill=blue!20, minimum width=1.2cm, minimum height=\blockHeight, rounded corners},
  encoderonly/.style={draw, fill=blue!20, minimum width=4.9cm, minimum height=1.0cm, rounded corners},
  encoderpart/.style={draw, fill=blue!30, minimum width=2.2cm, minimum height=0.6cm, rounded corners},
  decoder/.style={draw, fill=lime!30, minimum width=1.2cm, minimum height=\blockHeight, rounded corners},
  thresh/.style={draw, fill=red!20, minimum width=1.2cm, minimum height=\blockHeight, rounded corners},
  encprompts/.style={draw, fill=gray!30, minimum width=3.3cm, minimum height=1.2cm, rounded corners},
  promptenc/.style={draw, fill=purple!30, minimum width=3.4cm, minimum height=1.6cm, rounded corners},
  skip/.style={draw=black, thick, dashed, ->},
  conn/.style={->, thick},
  sum/.style={draw, circle, inner sep=0pt, minimum size=1em, fill=white, thick, font=\bfseries},
  >={Stealth[scale=0.7]}
}
\usepackage[breaklinks=true,colorlinks,bookmarks=false]{hyperref}
\usepackage{cleveref}
\pgfplotsset{compat=1.18}
\begin{document}
\title{Lite ENSAM: a lightweight cancer segmentation model for 3D Computed Tomography}
\titlerunning{Lite ENSAM}
%
\author{Agnar~Martin~Bjørnstad\orcidID{0009-0005-4207-6278} \and
Elias~Stenhede\orcidID{0009-0005-2654-4553} \and
Arian~Ranjbar\orcidID{0000-0002-0422-2255}
} 
\authorrunning{A.M. Bjørnstad et al.}
%
\institute{Faculty of Medicine, University of Oslo\\ Medical Technology \& E-health, Akershus University Hospital,\\ 1478 Lørenskog, Norway\\ \email{arian.ranjbar@medisin.uio.no}}
\maketitle              
\begin{abstract}

Accurate tumor size measurement is a cornerstone of evaluating cancer treatment response. The most widely adopted standard for this purpose is the Response Evaluation Criteria in Solid Tumors (RECIST) v1.1, which relies on measuring the longest tumor diameter in a single plane. However, volumetric measurements have been shown to provide a more reliable assessment of treatment effect. Their clinical adoption has been limited, though, due to the labor-intensive nature of manual volumetric annotation. In this paper, we present Lite ENSAM, a lightweight adaptation of the ENSAM architecture designed for efficient volumetric tumor segmentation from CT scans annotated with RECIST annotations. Lite ENSAM was submitted to the MICCAI FLARE 2025 Task 1: Pan-cancer Segmentation in CT Scans, Subtask 2, where it achieved a Dice Similarity Coefficient (DSC) of 60.7\% and a Normalized Surface Dice (NSD) of 63.6\% on the hidden test set, and an average total RAM time of 50.6\,GB\,s and an average inference time of 14.4\,s on CPU on the public validation dataset.

\end{abstract}

\section{Introduction}


Imaging plays a central role in assessing tumor response during cancer treatment, guiding oncologists in deciding whether to continue, adapt, or interrupt therapy. The most widely adopted framework for this purpose is RECIST v1.1~\cite{eisenhauer_new_2009}, valued for its simplicity, cost-effectiveness, and intuitiveness~\cite{fournier_twenty_2022}. Under RECIST, tumor size is quantified by measuring the longest diameter in a single plane of a scan.

While RECIST remains the clinical standard, volumetric tumor measurements have been shown to provide more sensitive and prognostic information. For example, volumetric analysis has been reported to better predict overall survival in lung cancer patients~\cite{hayes_comparison_2016} and to more accurately and earlier predict the pathological response to neoadjuvant chemotherapy in breast cancer than clinical assessment alone~\cite{hylton_locally_2012}. Despite these advantages, the adoption of volumetric measurements in clinical practice has been limited by the need for labor-intensive manual annotation of tumor volumes.

Advancements in computer vision offer a path forward by enabling automated volumetric lesion segmentation, thereby reducing annotation burden and potentially improving both the speed and accuracy of tumor response assessment~\cite{hering_improving_2024}. To advance this goal, the MICCAI FLARE 2025 challenge focuses on efficient and accurate medical image analysis under constrained computational resources. In particular, Task 1, Subtask 2 addresses the problem of converting RECIST annotations into full volumetric segmentations, bridging the gap between the widely used RECIST standard and clinically valuable volumetric measurements.




This task presents several challenges. Volumetric CT scans are inherently large, while inference must be performed under strict computational constraints. Models are limited to CPU execution with a maximum of 8 GB RAM, which requires highly memory- and compute-efficient designs. Furthermore, tumors vary widely in shape, size, and anatomical location, making robust generalization across diverse cases particularly difficult.

\subsection{Related work}




U‑Net–style architectures have long been dominant in medical image segmentation, consistently underpinning many top-performing models in major benchmarks. The original U‑Net~\cite{ronneberger_u-net_2015} demonstrated exceptional segmentation accuracy even with limited training data, and its encoder–decoder structure became a foundational paradigm. SegResNet \cite{myronenko_3d_2019}, a 3D U‑Net variant, was developed for brain tumor segmentation in multimodal MRI and notably secured first place in the BraTS 2018 challenge \cite{bakas_identifying_2019}. 
Similarly, nnU‑Net \cite{nnUNet} emerged as a highly influential, self‑configuring framework based on U‑Net. It automatically adapts architecture and training strategies based on dataset characteristics. In the 2021 Kidney Tumor Segmentation (KiTS21) challenge~\cite{heller_kits21_2023}, all top three solutions were U‑Net–based models using nnU‑Net. And in FLARE24 (whole-body cancer segmentation in CT scans), the winning solution employed a 3D U‑Net architecture~\cite{huang_efficient_2024}. Collectively, these findings display the effectiveness of U‑Net across imaging modalities.

ENSAM (Equivariant, Normalized, SAM in 3D)~\cite{stenhede_ensam_2025} is a foundation model for interactive 3D medical image segmentation. In the CVPR 2025: Foundation Models for Interactive 3D Biomedical Image Segmentation challenge\footnote{\url{https://www.codabench.org/competitions/7149/}}, ENSAM performed the best among models without external pretraining, outperforming the majority of baseline methods. ENSAM adopts a 3D U-Net-inspired design based on SegResNet~\cite{myronenko20183d}.

\subsection{Objective and contribution}
In this paper, we adapt the ENSAM architecture for the task of tumor segmentation in 3D CT images. In particular, we further improve performance and introduce a lightweight variant, named Lite ENSAM, optimized for efficient inference on CPU.


\section{Method}
    
\begin{figure}[htbp]
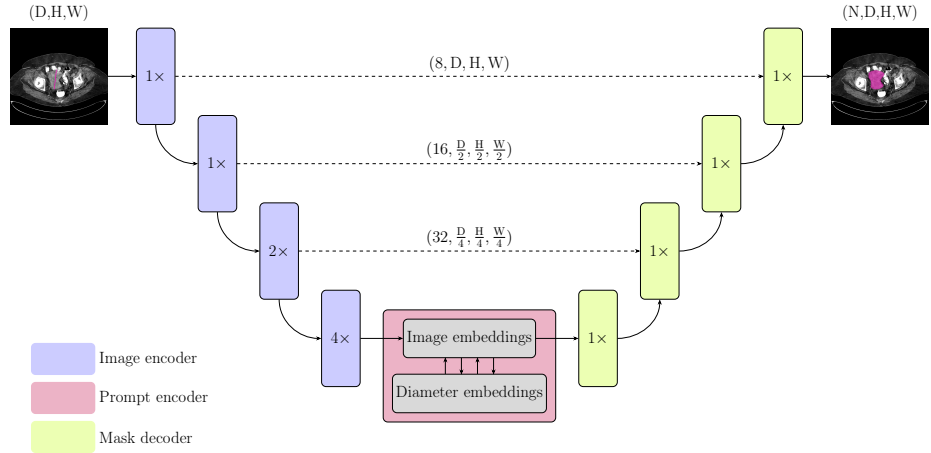

      \includestandalone[width=\textwidth]{tikz/model}
      \caption{Network architecture of Lite ENSAM, consisting of three main components: image encoder, prompt encoder, and mask decoder. The diameter markings are incorporated via cross-attention between image embeddings and diameter embeddings in the bottom part of the U-Net.
      }\titlerunning{ENSAM}
      \label{fig:Network}
    \end{figure}

    Lite ENSAM is architecturally similar to ENSAM, consisting of a 3D U-Net, augmented with a SAM-style attention mechanism~\cite{SAM-ICCV23} linking the network's deepest layer to the embeddings of RECIST annotated tumors. The attention mechanism is implemented using a normalized transformer architecture~\cite{loshchilov2025ngpt} with Lie Rotational Positional Encoding (LieRE)~\cite{ostmeier_liere_20252}.  An overview of the architecture is shown in \Cref{fig:Network}, and the following section provides a brief description of the model.
    
\subsection{Network architecture}



\subsubsection{Image encoder}
\label{subsubsec:image_encoder}
    The image encoder is composed of multiple blocks that operate at progressively lower resolutions. Each block consists of an instance normalization layer, followed by two repetitions of a 3D convolutional layer and ReLU activation, and a skip connection. Downsampling between blocks is performed using a convolutional layer with a stride 2.

\subsubsection{Encoding of RECIST annotation}
\label{subsubsection:encoding_of_recist_annotation}
    

    In the MICCAI FLARE 2025 Task 1, Subtask 2, tumors are annotated according to RECIST using their longest in-plane diameter on the axial (z) slice with the largest cross-sectional area. Representing RECIST annotations as dense volumetric masks would be both information-sparse and memory-intensive, since the marking itself provides no spatial context beyond its two endpoints. Instead, we extract and encode only the diameter endpoints, which contain all available localization information.

    The interaction between the diameter and image embeddings takes place at the bottleneck of the U-Net, where cross-attention with Lie Rotational Positional Encoding is applied. For details of the cross-attention mechanism and relative position encoding, we refer to the ENSAM paper \cite{stenhede_ensam_2025}.

\subsubsection{Image decoder}
    The mask decoder mirrors the image encoder, but employs a single residual block at each upsampling stage. Skip-connection activations are concatenated along the channel dimension and processed by a 3D residual block, followed by trilinear upsampling. The final layer outputs logits with the same spatial dimensions as the input, representing the predicted segmentation mask. Voxels with logits greater than zero are classified as tumor.

\subsubsection{Modifications to ENSAM}
    To adapt the ENSAM model for this competition, several modifications were required. First, the parsing and processing pipelines had to be updated. ENSAM was originally designed for interactive medical image segmentation, where it received a bounding box or user click for each object of interest to indicate its location. Based on these inputs, the model produced object segmentations and subsequently accepted refinement clicks, each pointing to the region of largest error for a given object. This interactive process allowed the model to iteratively improve its segmentation. In contrast, the current competition provides CT volumes together with diameter markings for each tumor. Consequently, ENSAM was modified to accept diameter points as an input modality and to operate in a non-interactive setting, with the refinement stage removed. 

    As inference is performed on a CPU with a maximum of 8 GB RAM, the model must operate under substantially more restrictive memory and computational constraints. To accommodate this, Lite ENSAM was designed to be more efficient in both memory footprint and compute. Specifically, the number of output filters in the image decoder and the embedding dimension were reduced by half, and the maximum patch size of the input volume to the decoder was limited to one-quarter of that used in the original ENSAM.

    
    
    Finally, the post-processing step was re-implemented to be more memory- and compute-efficient, which was essential since the original implementation exceeded the available memory for certain inputs. To address this, the number of full-volume allocations was reduced by restructuring the computation of intermediate tensors, thereby reducing both memory footprint and computational overhead. Additionally, to verify the presence of all classes in the final prediction, we replaced torch.unique with torch.bincount, which avoids sorting and is therefore faster. On CPU, these optimizations yielded a $30\times$ speedup of the post-processing stage on the validation set, reducing its average runtime to 0.2 seconds on the hardware specified in \Cref{table:env}.

\subsection{Pre-processing}

    All volumes were min-max normalized to range between 0 and 1 during training and inference. Besides normalization, preprocessing was limited to cropping, resizing and flipping volumes, ensuring training can run within the specified VRAM limits. A maximum size of $128 \times 128 \times 128$ was used during training and inference. Anisotropy was not explicitly considered, and unlabelled or pseudolabelled data were not used for training.



\subsection{Post-processing}
    Each volume may contain multiple annotated tumors, all of which are included in the ground truth segmentation. To reduce the impact of low-confidence predictions for individual tumors, the model is required to predict all annotated tumors in the final output. Specifically, if all logits for a given tumor are below zero, the logits within the sphere defined by its RECIST diameter are incrementally increased by an exponential offset until at least one voxel is classified as belonging to that tumor.

\section{Experiments}
\subsection{Dataset and evaluation measures}
Data was provided by the MICCAI FLARE 2025 challenge, and the segmentation targets cover various lesions. The training dataset is curated from more than 50 medical centers under the license permission, including TCIA~\cite{TCIA}, LiTS~\cite{LiTS}, MSD~\cite{simpson2019MSD}, KiTS~\cite{KiTS19,KiTSDataset,KiTS21}, autoPET~\cite{autoPET-Data,autoPET-MICCAI22}, TotalSegmentator~\cite{TotalSegmentator}, and AbdomenCT-1K~\cite{AbdomenCT-1K}, FLARE 2023~\cite{FLARE23}, DeepLesion~\cite{deeplesion}, COVID-19-CT-Seg-Benchmark~\cite{COVID-19-CT-Seg-Benchmark}, COVID-19-20~\cite{COVID-19-20}, CHOS~\cite{CHAOS2021}, LNDB~\cite{LNDB}, and LIDC~\cite{LIDC}. The training set includes more than 10,000 abdominal CT scans, where 2,200 CT scans with partial labels and 1,800 CT scans are without labels. The validation and testing sets include 100 and 400 CT scans, respectively, which cover various abdominal cancer types, such as liver cancer, kidney cancer, pancreas cancer, colon cancer, gastric cancer, and so on. The lesion annotation process used ITK-SNAP~\cite{ITKSNAP}, nnU-Net~\cite{nnUNet}, MedSAM~\cite{MedSAM,MedSAM2}, and Slicer Plugins~\cite{Slicer,MedSAM2}.

The evaluation metrics encompass two accuracy measures: DSC and NSD, alongside two efficiency measures: running time and area under the CPU memory-time curve. These metrics collectively contribute to the ranking computation. Furthermore, the running time and CPU memory consumption are considered within tolerances of 45 seconds and 4 GB, respectively.

\subsection{Implementation details}
\subsubsection{Environment settings}
The development environments and requirements are presented in \Cref{table:env}.

\begin{table}[!htbp]
    \caption{Development environments and hardware.}
    \label{table:env}
    \centering
    \begin{tabular}{@{}ll@{}}
        \toprule
        Component                   & Specification \\
        \midrule
        System                      & Debian 12 \\
        CPU                         & Intel(R) Core(TM) i9-14900KF \\
        RAM                         & 2$\times$48\,GB; 4800\,MT/s \\
        GPU                         & NVIDIA GeForce RTX 5090 32\,GB \\
        CUDA version                & 12.8 \\
        Programming language        & Python 3.12 \\
        Deep learning framework     & PyTorch 2.7.0, Torchvision 0.22.0 \\
        \bottomrule
    \end{tabular}
\end{table}

\subsubsection{Training protocols}


The provided dataset\footnote{\url{https://huggingface.co/datasets/FLARE-MedFM/FLARE-Task1-PancancerRECIST-to-3D}} was used exclusively for model training and model selection; no partial- or unlabeled data were used. The samples were categorized by source and uniformly sampled across categories to mitigate class imbalance. To concentrate computational resources on relevant structures, training volumes were randomly cropped around annotated regions with a variable margin of 1–64 voxels. Volumes exceeding a predefined size threshold were downscaled via max pooling to fit within GPU memory. Given the variability in sample dimensions, the batch size was fixed at 1. To ensure compatibility with the network architecture, zero-padding was applied so that all dimensions were divisible by 8. Each axis was randomly flipped with a probability of 50\%. An overview of training configuration is provided in \Cref{table:training_protocols}. Consistent with ENSAM, Lite ENSAM was trained using a single GPU.

%
%
%
%

\begin{table}[H]
    \caption{Training protocols.}
    \label{table:training_protocols}
    \centering
    \begin{tabular}{@{}ll@{}}
        \toprule
        Parameter                   & Specification \\
        \midrule
        Batch size                      &  1 \\
        Patch size                      & Variable \\
        Total epochs                    & 75 \\ 
        Optimizer                       &  Muon and AdamW\\
        Initial learning rate (lr)      &  0.002\\
        Training time                   &  35h \\
        Loss function                   & Soft Dice + $2 \cdot$BCE \\
        Number of model parameters      & 1.3M\\ 
        Number of flops                 & 29G \\ 
        CO$_2$eq                        & 0.214 Kg \\ 
        \bottomrule
    \end{tabular}
\end{table}

\section{Results and discussion}
%
%
%
%


\begin{table}[htbp]
\caption{Quantitative evaluation results.}
\label{tab:final-results}
\centering
\newcolumntype{C}{>{\centering\arraybackslash}p{1.8cm}} 
\begin{tabular}{lCCCC}
\toprule
& \multicolumn{2}{c}{Public Validation} & \multicolumn{2}{c}{Testing} \\
\cmidrule(lr){2-3} \cmidrule(lr){4-5}
& DSC (\%) & NSD (\%) & DSC (\%)& NSD (\%)\\
\midrule
Lite ENSAM & 76.1 $\pm$ 16.3 & 78.9 $\pm$ 19.1 & 60.7 & 63.6 \\
\bottomrule
\end{tabular}
\end{table}


\subsection{Quantitative results on validation set}
%

Quantitative results are shown in \Cref{tab:final-results}. On the public validation dataset, our method achieved a DSC of 76.1\% and an NSD of 78.9\%.

\begin{table}[!htbp]
    \caption{Quantitative evaluation of segmentation efficiency in terms of the running time and RAM use on the validation set. Total RAM use denotes the area under the RAM-Time curve. 
    }
    \label{table:efficiency}
    \centering
    \begin{tabular}{@{}lllll@{}}
        \toprule
        Case ID & Image Size      & Running Time (s) & Max RAM (GB) & Total RAM (GB s) \\
        \midrule
        
        0001	&  (55, 512, 512)  & 17.2  & 4.0 & 55.9   \\
        0011	&  (100, 512, 512) & 14.8   & 4.8	& 52.1   \\
        0021	&  (152, 512, 512) & 14.3  & 5.0	& 51.0   \\
        0031	&  (215, 512, 512) & 13.6  & 4.7	& 47.4   \\
        0041	&  (157, 512, 512) & 14.3  & 4.7	& 50.0   \\
        0051	&  (171, 512, 512) & 14.2  & 4.4	& 49.8   \\
        0061	&  (147, 512, 512) & 14.5  & 5.2	& 52.7   \\
        0071	&  (240, 512, 512) & 15.0  & 5.6	& 54.7   \\
        0081	&  (103, 512, 512) & 13.2  & 4.6	& 46.3   \\
        0091	&  (137, 512, 512) & 13.0  & 4.5 & 45.5   \\ \midrule
        Average &                  & 14.4  & 4.7 & 50.6   \\
        \bottomrule
    \end{tabular}
\end{table}


\subsection{Qualitative results on validation set}



\begin{figure}[htbp]
    \centering
    \resizebox{\textwidth}{!}{\input{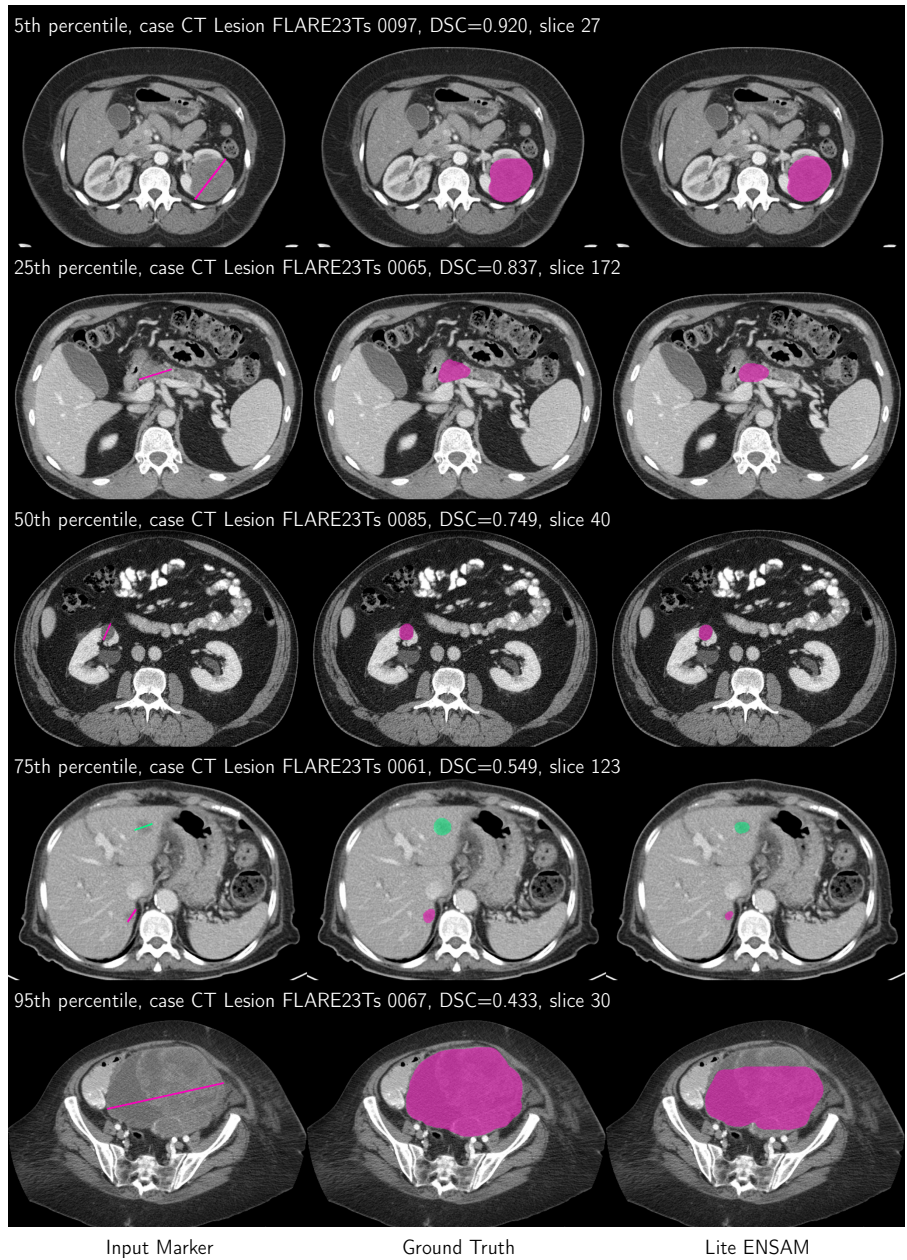}}
    \caption{Example slices from five volumes in the validation set. The volumes were chosen based on their Dice scores, corresponding to the 5th, 25th, 50th, 75th, and 95th percentiles. For each volume, the slice aligned with an input marker from class 1 was selected.}
    \label{fig:overlay_plot}
\end{figure}

\Cref{fig:overlay_plot} presents predictions for five representative validation cases, corresponding to the 5th, 25th, 50th, 75th, and 95th percentiles of the Dice score relative to the ground truth. The slice indices are selected to match the plane of the first RECIST marker. In the cases with the lowest Dice score, the model seems to undersegment the tumors.


%
%
%
%

\subsection{Segmentation efficiency results on validation set}

\Cref{table:efficiency} shows the inference efficiency on a selection of validation cases. For all the cases, there is a significant margin to the maximum RAM threshold of 8GB. Inference on healthy CT scans were not performed as they do not have any diameter markings.


\subsection{Limitation and future work}



While Lite ENSAM demonstrates competitive performance under strict resource constraints, several limitations remain that suggest directions for future work. First, the model was trained with constrained compute and exclusively on the provided training data. Training on a larger and more diverse dataset would likely improve generalization and segmentation accuracy. Second, the impact of additional data augmentation other than axis flips was not explored. Such augmentations could further enhance robustness to variations in tumor appearance. Finally, to strengthen its clinical relevance, future work should extend evaluation beyond CT to other imaging modalities widely used in treatment response assessment, such as Magnetic Resonance Imaging. This would increase the model’s applicability in clinical practice and enable validation of its performance across modalities.



\section{Conclusion}


In this paper, we introduced Lite ENSAM, a lightweight version of the ENSAM architecture for tumor segmentation in 3D CT scans with RECIST annotations. Lite ENSAM was specifically developed to enable inference on CPU under a strict memory constraint of 8 GB RAM. On the public validation dataset, it achieved a DSC of 76.1\% and an NSD of 79.0\%. This work demonstrates that the ENSAM architecture can be effectively adapted to low-compute, low-memory environments while maintaining competitive segmentation performance, and further validates its effectiveness when trained with diameter-based annotations.

\subsubsection{Acknowledgements} The authors of this paper declare that the segmentation method they implemented for participation in the FLARE 2025 challenge has not used any pre-trained models nor additional datasets other than those provided by the organizers. The proposed solution is fully automatic without any manual intervention. We thank all data owners for making the CT scans publicly available and CodaLab~\cite{codabench} for hosting the challenge platform. The authors express their appreciation to Novartis Norge AS and Akershus University Hospital for funding this work.


\section*{Disclosure of Interests} The authors declare no competing interests relevant to this work.


%
%
\printbibliography

\newpage

\end{document}